# Linguistic Analysis of Sinhala YouTube Comments on Sinhala Music Videos: A Dataset Study

W. M. Yomal De Mel, Nisansa de Silva

Department of Computer Science & Engineering, University of Moratuwa, Moratuwa, Sri Lanka

{mario.23,NisansaDdS}@cse.mrt.ac.lk

*Abstract*—This research investigates the area of Music Information Retrieval (MIR) and Music Emotion Recognition (MER) in relation to Sinhala songs, an underexplored field in music studies. The purpose of this study is to analyze the behavior of Sinhala comments on YouTube Sinhala song videos using social media comments as primary data sources. These included comments from 27 YouTube videos containing 20 different Sinhala songs, which were carefully selected so that strict linguistic reliability would be maintained and relevancy ensured. This process led to a total of 93,116 comments being gathered upon which the dataset was refined further by advanced filtering methods and transliteration mechanisms resulting into 63,471 Sinhala comments. Additionally, 964 stop-words specific for the Sinhala language were algorithmically derived out of which 182 matched exactly with English stop-words from NLTK corpus once translated. Also, comparisons were made between general domain corpora in Sinhala against the YouTube Comment Corpus in Sinhala confirming latter as good representation of general domain. The meticulously curated data set as well as the derived stop-words form important resources for future research in the fields of MIR and MER, since they could be used and demonstrate that there are possibilities with computational techniques to solve complex musical experiences across varied cultural traditions

## I. Introduction

Music Information Retrieval (MIR) landscape is going through a period of transformation with an increasing interest in uncovering emotional dimensions within different musical traditions. This research embarks on a scholarly exploration situated at the intersection of MIR and Music Emotion Recognition (MER) specific to Sinhala songs, an underexplored territory within the broader spectrum of music research. The scarcity of Natural Language Processing (NLP) resources for the Sinhala language makes assessing the emotional impact of Sinhala music even more difficult. While extensive research has been conducted on analyzing social media comments for English songs, the unique linguistic and cultural characteristics of Sinhala present distinctive challenges and opportunities for understanding music emotions.

Sri Lankan music, which is distinguished by its varied musical genres and rich cultural legacy, is evidence of the artistic diversity embedded in the traditions of the nation. Despite its cultural significance, the scholarly exploration of Sri Lankan music, particularly from the perspective of MIR, remains limited. By examining the linguistic patterns and usage in social media comments on Sinhala songs, this study aims to bridge this gap. Researchers aim to gain an understanding of the lexical diversity in Sinhala music by looking at these observations.

The primary objectives of this research include the development of a comprehensive dataset[1] of Sinhala YouTube comments on Sinhala songs, the identification of frequently occurring words and word pairs, and a comparison of these linguistic elements with those found in general Sinhala domain corpora. This structured approach integrates both linguistic and computational methodologies to navigate the complex linguistic landscape of Sinhala songs. Additionally, the utilization of social media comments as primary data sources serves as a unique lens into the spontaneous and authentic reactions of listeners, enriching the understanding of linguistic patterns within the Sinhala cultural domain.

This study emphasizes the importance of identifying and analyzing the usage of Sinhala words in YouTube comments compared to their usage in the general domain. Understanding these differences can help highlight the richness of the Sinhala YouTube comment domain, identify rarely used words within YouTube comments that are common in the general domain, and locate frequently used words within YouTube comments that are not as common in the general domain. The authors utilized Sinhala Wikipedia articles, Sinhala newspaper articles, and data from previous research studies on Sinhala NLP which includes data from government documents, ensuring the inclusion of the majority of freely available Sinhala corpora for NLP analysis. This robust dataset is crucial for a thorough comparative analysis, enabling the authors to draw meaningful insights into the distinct characteristics of word usage in Sinhala YouTube comments versus the general Sinhala language domain.

The thoroughness of the preprocessing process ensures the integrity and reliability of the dataset, allowing for meaningful insights to be derived from the preceding analysis. It also focuses on specific Sinhala language characters, like the zero-width joiner, when filtering data, making sure that language-specific factors are properly considered. The dataset presented in this research paper offers a valuable resource for investigating the linguistic dimensions of Sinhala music within the domain of social media discourse. Its thorough preprocessing and extensive coverage of a wide range of songs, eras, and emotional expressions provide the foundation for insightful understanding of the complex interplay between language, emotion, and music in the Sinhala cultural context.

---

[1]https://github.com/Yomald93/Linguistic-Analysis-of-Sinhala-YouTube-Comments-on-Sinhala-Music-Videos.git

## II. Related work

### A. Sinhala Corpus Creation and Evaluation

For any language, the foundation of effective NLP applications is the availability of robust corpora. In the case of Sinhala, notable efforts have been made to create such resources. by web crawling, incorporating Jathaka stories, web crawlin news paper articles, extracting content from government documents etc as per de Silva [1]. An intial contribution to develop a Sinhala corpus was given by [2] by collected by online resources including news, academic, creative writing, spoken and gazette. Then de Silva [3] has created a data set from Sinhala news posts. Further a Sinhala-English parallel dataset created by [4]. Thereafter a significant contibution for Sinhala corpus creation was made by Wijeratne and de Silva [5] by making freely available a massive dataset using Sinhala facebook posts. It was also recognised that a Sinhala corpus created by Dhananjaya et al. [6], which they have claimed as the largest monolingual Sinhala corpus at the time of their publication has also provided a reasonable contribution for Sinhala corpus creation. However, these corpora are relatively modest when compared to the vast English corpora available. Recently, a significant leap was made with the creation of a corpus consisting of 506,932 Sinhala news articles, marking a substantial improvement in both size and coverage over previously developed Sinhala corpora [7].

### B. Music Information Retrieval (MIR)

Music Information Retrieval (MIR) refers to strategies and techniques aimed at enabling efficient access to music collections, both new and historical. This field addresses the growing expectations for search and browse functionality, and has gathered significant attention from academic and industrial research entities, as well as libraries and archives. MIR benefits three main groups: industry professionals involved in music recording and distribution, end users seeking personalized music experiences, and professionals such as performers, teachers, musicologists, and producers [8]. Various music representations have been tested in MIR, including pitch histograms, the Generalized Pitch Interval Representation (GPIR), Spiral Arrays, and specialized graph structures [9].

### C. Music Emotion Recognition (MER)

Music Emotion Recognition (MER) is a subfield of Music Information Retrieval (MIR) that employs computational methods to automatically identify and analyze the emotional content in music [10]. The primary goal of MER is to develop algorithms and models capable of extracting relevant features from songs and using machine learning techniques to classify and understand their emotional characteristics. This endeavour has significant implications for various applications, including music recommendation systems, mood-based playlist generation, and affective computing [11].

Researchers in MER face numerous challenges due to the subjective nature of emotional perception and the absence of a standardized emotion taxonomy [12]. Additionally, the complex nature of music and emotions requires robust feature extraction and classification methodologies. Despite these challenges, the field of MER is rapidly evolving, driven by its potential to deepen our understanding of the emotional aspects of music and enhance user experiences.

Music has a profound impact on human emotions and behaviour, activating brain regions involved in reward, motivation, and emotion processing [13]. Moreover, music can elicit a diverse range of emotional responses through its structural features, such as melody, harmony, rhythm, and timbre [14]. This dynamic interaction between music and emotions presents opportunities for personalized music recommendation systems, affective computing, and therapeutic interventions [9].

The historical and theoretical exploration of the relationship between music and emotions has laid the groundwork for MER. From ancient beliefs in the emotive power of music to modern-day interdisciplinary research, scholars have developed various emotion models and taxonomies to understand this complex relationship [12]. Further research in MER is being driven by the growing desire for automated systems to categorize and retrieve music based on its emotional content, which is being brought about by the widespread consumption of digital music.

Despite the progress made in MER, challenges such as acquiring reliable emotion annotations and capturing temporal aspects of music emotion persist [15]. Addressing these challenges requires a multidisciplinary approach, drawing expertise from fields such as signal processing, machine learning, psychology, and musicology. By advancing our understanding and development of MER systems, researchers can unlock new insights into the sophisticated relationship between musical features and emotional responses, with implications for psychology, cognitive science, and music theory [11].

### D. Word Embedding

With the advancements in the feild of neural networks have significantly broadened the scope of research into various approaches for deriving word embeddings [16]. Word embeddings are real-valued vector representations of words that capture both semantic and syntactic meanings from large, unlabeled corpora. This technique has become a powerful tool in modern Natural Language Processing (NLP) tasks, including semantic analysis [17].

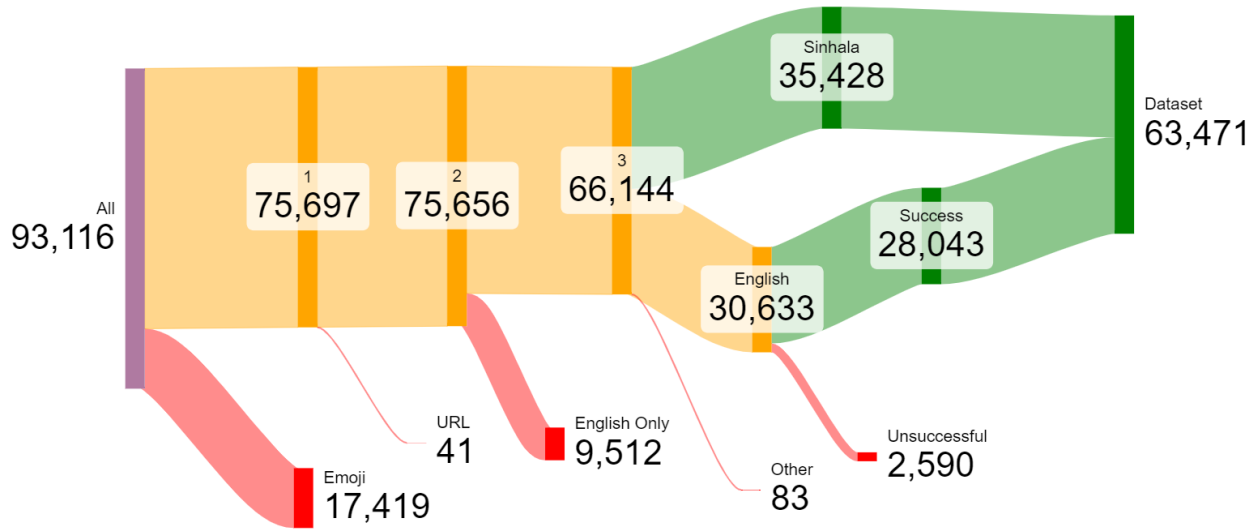

Fig. 1: Stages of the Data Preparation Process

Word2Vec developed by Mikolov et al. [18], represents a breakthrough in neural word embedding models. It is trained using a straightforward feed-forward neural network, which has made it highly influential in the domain of NLP due to its simplicity and effectiveness. Word2Vec excels in constructing dense, low-dimensional vector representations of words, which has been crucial in advancing various NLP applications [19]. The Word2Vec model comes in two variants: the continuous bag of words (CBOW) model and the continuous skip-gram model. Both variants are designed to capture the semantic and syntactic relationships between words, despite ignoring the word order information. The CBOW model predicts the target word from the context words, while the skip-gram model does the reverse by predicting context words from a given target word. The ability of these models to encapsulate meaningful linguistic patterns has been demonstrated in numerous studies, showing their competency in understanding word relationships.

## III. Data Preparation

Data collection for this study involved gathering comments from 27 YouTube videos featuring 20 Sinhala songs spanning various eras of Sinhala music. Due to the unavailability of certain songs on the original artist's YouTube channel, multiple versions performed by the original artist were considered. The initial step in data retrieval utilized Google App Script, which facilitated the extraction of comments by providing the YouTube video ID to an online macro. In total, 93,116 comments were obtained from this process, forming the basis for subsequent analysis. The subsequent filtration stages of the data preparation process described in this section are illustrated in Figure 1.

The collected comments underwent extensive preprocessing to prepare the data for analysis. This involved addressing various challenges inherent in online comment datasets, such as the presence of emojis, URLs, numbers, dates, symbols [20–22]. To ensure the quality and reliability of the dataset, a further rigorous data-cleaning process was implemented. This also included removing non-Sinhala language comments, eliminating irrelevant or spam-like content, and standardizing the format of the remaining comments to facilitate analysis [23]. By undertaking these preprocessing steps, the dataset was brought to an analyzable level, ready for in-depth exploration of emotional dimensions within Sinhala music discourse.

Utilizing the aforementioned methodology, a subset comprising 75,656 data rows was identified for subsequent analysis. This subset underwent comparison against the nltk Corpus of English words to distinguish comments exclusively composed of English vocabulary. This comparative analysis facilitated the filtration of 9,512 data rows, characterized by solely English words. The remaining 66,144 data rows were then examined in further detail to check whether any Sinhala characters were included in the comments. Through this process, 35,428 comments featuring Sinhala characters were identified and subsequently incorporated into the final dataset. The validation protocol encompassed thorough verification of each comment vis-à-vis the Unicode range stipulated for Sinhala characters, thereby ensuring linguistic fidelity. Following this classification, it was determined that the remaining 30,716 comments had a combination of non-English and English characters, increasing the dataset's linguistic richness and analytical potential. This methodological rigour emphasizes how important it is to maintain data integrity and relevance within scholarly pursuits.

In the subsequent stages of data processing, the dataset underwent meticulous segregation based on linguistic characteristics, resulting in the identification of 83 comments containing non-English characters and 30,633 comments composed solely of English characters. Following this classification, a transliteration process was initiated employing the Google Transliterator API.

The objective of the transliteration procedure was to convert the 30,633 English character comments into Sinhala script. Leveraging the Google Transliterator API, efforts were made to transliterate the English text into Sinhala. Remarkably, 28,043 comments were successfully transliterated, resulting in the creation of a final dataset with 63,471 Sinhala comments.

The resultant dataset, post-transliteration, encapsulated 63,471 Sinhala comments, constituting approximately 68% of the initial dataset. This consolidation of Sinhala comments represents a substantial proportion of the original dataset, confirming the data's representativeness and robustness for subsequent analytical endeavours.

Table I presents an overview of the dataset's composition, detailing the contributions from various songs, which collectively form a corpus of 63,471 comments. Among the songs, "අන් සතු ඔබ" emerges with the highest number of Sinhala comments, totalling 14,100, underscoring its significance within the dataset. Conversely, "පියාණනි" records the lowest count of comments at 130, highlighting a notable variation in engagement across different songs. This disparity in comment counts reflects the varying levels of audience interaction and interest elicited by each song, providing valuable insights into the dataset's composition and the audience's preferences.

TABLE I: Comment Counts by Artist and Song

| Artist | Song | Comment Count |
|---|---|---|
| Thisara Weerasinghe | අන් සතු ඔබ | 14100 |
| Sarith and Surith | සල්ලි සල්ලි | 12254 |
| Dinesh Gamage | දැනෙනා තුරු මා | 5378 |
| Methun SK | කාරි නෑ සඳ | 3935 |
| Saman Lenin | අමුඩේ | 3757 |
| BnS | උන්මාද ප්‍රේම ගීය | 3292 |
| Danith Sri | එහෙම දේවල් නෑ හිතේ | 2825 |
| Sanuka Wick | පෙරවදනක් | 2634 |
| Raini Charuka | කළුවරට හිත බය | 2596 |
| Ridma | සොබනා | 2058 |
| Prageeth Perera | කෝමලියා | 1871 |
| Abisheka and Mihdu | දන්නවාද ආදරේ නීතිය | 1722 |
| Sanuka Wick | සරාගයේ | 1506 |
| Sandeep Jayalath | නුරාවී | 1372 |
| Sanuka | මෝහිනී | 1319 |
| Saman Lenin | අම්බරුවෝ | 809 |
| Victor Rathnayake | තනිවෙන්නට මගේ ලොවේ | 791 |
| Nanda Malini | මා සඳට කැමති බව | 735 |
| Victor Rathnayake | අපෙ හැඟුම් වලට | 387 |
| Ashanthi | පියාණනි | 130 |

In summary, the data collection and preprocessing stages established crucial preparatory steps in this research endeavour. Through systematic extraction and thorough cleaning of comments from YouTube videos, a comprehensive dataset was compiled, laying the groundwork for subsequent analysis of emotional responses elicited by Sinhala songs across different eras. The thoroughness of the preprocessing process ensures the integrity and reliability of the dataset, allowing for meaningful insights to be gleaned from the ensuing analysis.

This methodological approach underscores the significance of systematic data processing techniques in ensuring the integrity and relevance of research outcomes, particularly in multilingual domains. The utilization of the Google Transliterator API has proven instrumental in effectively expanding the linguistic diversity of the dataset, thereby facilitating a more comprehensive analysis of Sinhala language content. The implementation of thorough data processing procedures is crucial in promoting robust and nuanced research in linguistically diverse contexts.

## IV. Data Analysis

In this section, an in-depth analysis of the finalized dataset reveals previously undiscovered insights within the comments. Through rigorous scrutiny, hidden data patterns and nuanced aspects of the comment content emerge, enriching the understanding of the dataset's underlying dynamics. This careful investigation not only reveals knowledge that was previously unknown, but it also emphasizes how crucial careful analysis is to revealing hidden discoveries for academic research. The entire data analysis process is shown in the Figure 2.

Within the curated comments, a vast array of 67,959 unique words was uncovered. These words were identified as clusters of characters separated by one or more spaces. When separating these words it was crucial to pay attention on the fact that there are some Sinhala words that have the zero-width-joiner character.

Initially, the focus was on these character clusters. However, a meticulous filtration process ensued to refine the dataset. This process targeted character clusters consisting solely of numbers, those containing a blend of numbers and letters, emojis, or characters not belonging to the Sinhala language. Following this refinement, the count of unique Sinhala words swelled to 54,834. This methodical approach not only increased the comprehensiveness of the dataset but also facilitated a more focused analysis of linguistically relevant content.

Then the existence of these unique words on the refined comments was analysed. It was noted that a higher number of comments consist of two unique Sinhala words. The highest number of unique words identified on a comment was 639. The distribution of the above-mentioned unique words was arranged such that 25% of the comments include one unique word, 50% of the comments include less than four unique words and 75% of the comments less than nineteen unique words. Figure 3 visualises the distribution of the number of comments against the unique word counts which is the comment length.

The examination of single-word frequencies and word pair frequencies within the comments sheds light on the distribution of linguistic elements, revealing nuanced patterns within the dataset. Notably, an analysis of word pairs underscores the richness of linguistic expression encapsulated within the Sinhala language, with an extensive collection of unique word pairings comprising the discourse. This analysis revealed a vast array of 328,922 distinct word pairs, indicative of the diverse linguistic landscape inherent in the comments. Table II presents the top ten frequencies for both unique Sinhala words and word pairs, offering insight into the prevalent themes and sentiments expressed within the comment section.

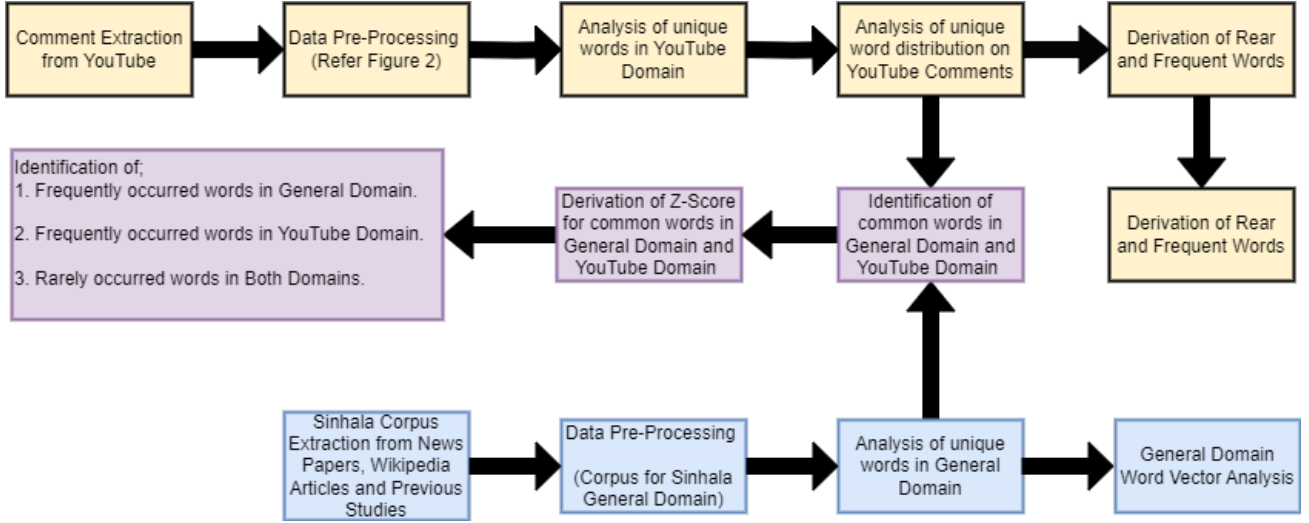

Fig. 2: Data Analysis Process

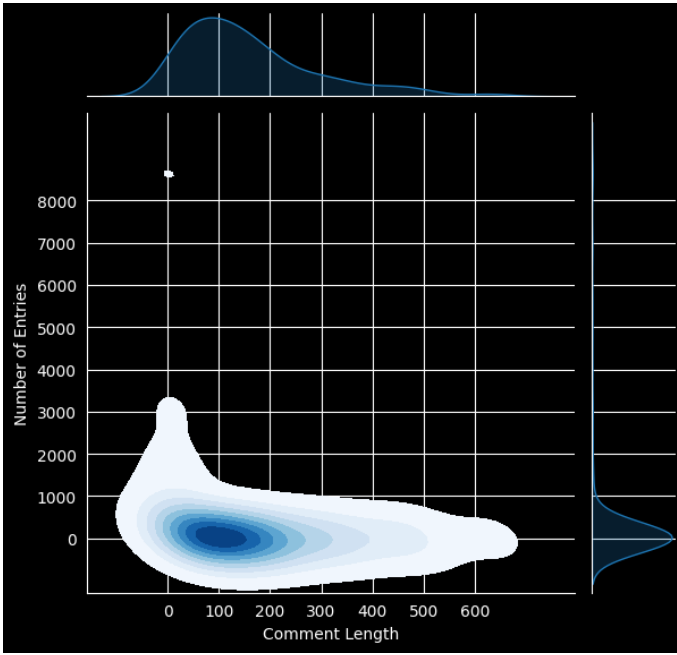

Fig. 3: Distribution of Comment Length

Additionally, the conclusions drawn from the highest frequencies highlight common themes and opinions expressed by commentators, highlighting certain terms and word combinations that frequently appear in the conversation. The prominence of terms such as "ලස්සනයි", "සුපිරි", and word pairs like "කියන්න වචන", "කොච්චර ඇහුවත්" and "හරිම ලස්සනයි" signifies an overarching appreciation for the aesthetic appeal and emotional resonance of the songs under discussion. These lexical choices not only reflect the linguistic preferences of commenters but also offer glimpses into the cultural and emotional responses evoked by the music. Thus, the examination of word frequencies and pairings serves as a window into the complex interactions of audience engagement, revealing the intricate interplay between language, emotion, and cultural domain within the context of online discourse.

TABLE II: Top 10 Word and Word Pair Frequencies

| Word | Frequency | Word Pair | Frequency |
|---|---|---|---|
| මේ | 6836 | මේ වගේ | 947 |
| එක | 5989 | කියන්න වචන | 799 |
| ඒ | 5007 | හරිම ලස්සනයි | 699 |
| නෑ | 4451 | කොච්චර ඇහුවත් | 699 |
| ලස්සනයි | 3236 | මේ සින්දුව | 648 |
| වගේ | 3139 | ඇහුවත් එපා | 586 |
| සෝන්ග් | 2907 | කාරි නෑ | 569 |
| එකක් | 2851 | ඔබ වෙත | 559 |
| නම් | 2821 | වචන නෑ | 550 |
| සුපිරි | 2748 | සෝන්ග් එක | 504 |

### A. Derivation of Stop Words

In the domain of textual analysis, the identification of stop words holds greater importance as they typically noticeable as global elements across diverse textual compositions, contributing minimal informational value. These stop words, by their inherent nature, permeate nearly every sentence, exerting a marginal impact on the overall semantic interpretation. In the domain of sentiment analysis for Sinhala YouTube comments, determining these stop words assumes particular significance, as they could potentially skew the sentiment analysis outcomes. The identification and subsequent removal of such stop words serve as a foundational step in refining the accuracy and reliability of sentiment analysis algorithms applied to Sinhala textual data.

To facilitate the identification of stop words within the Sinhala YouTube comment corpus, a method similar to that proposed by Wijeratne and de Silva [5]. was adopted. Initially, an exhaustive collection of word frequencies was combined from the entire corpus, mirroring the process undertaken in previous linguistic analyses.

Statistical measures such as the standard deviation ($\sigma$) and the mean ($\mu$) were computed for the remaining word frequencies. Leveraging these corpus-specific statistical parameters, the frequencies of individual words were standardized utilizing a well-established equation, wherein each word's frequency was transformed into a standardized score denoted as 'z.' This normalization process facilitated a comparative assessment of word frequencies across the corpus, enabling the way for the identification of outliers at the upper end of the distribution – a critical step in pinpointing potential stop words.

$$z = \frac{(x - \mu)}{\sigma} \quad (1)$$

Ultimately, words with standardized frequencies surpassing a predefined threshold – in this case, a threshold set at 3.0 – were designated as frequently occurring words in the YouTube Sinhala comment domain for Sinhala songs. The distribution of the Z value for word frequencies is shown on Figure 4a and Figure 4b show the magnified version of the frequency graph focusing the range from -3.5 to 3.5. The rationale behind this threshold mechanism lies in its alignment with established norms and methodologies prevalent in linguistic analyses, ensuring a harmonious integration of the stop-word identification process within the broader framework of textual analysis. The completing these methodological steps curated list of frequently occurring words tailored to the unique characteristics of Sinhala YouTube comments was produced, thereby furnishing researchers with a vital resource to enhance the precision and efficacy of sentiment analysis endeavours in Sinhala textual data.

From this process, the authors were able to identify the list of 964 frequently occurring words. This list of words was translated to English to check the availability on the English stop word list on nltk corpus. This was done in order to identify the list of true stop words out of the frequently occurred words in the Sinhala YouTube comment domain for Sinhala songs. Out of those 182 words were available on the nltk stop word list. Further above identified list of frequently occurred words were compared with the list of Sinhala stop words proposed by Lakmal et al. [24] which included 191 stop words. Out above 964 frequently occurred words 53 were included in the aforementioned Sinhala stop word list.

## B. Comparison of Sinhala YouTube Comment Corpus for Sinhala Songs with Sinhala General Domain

This study emphasizes the importance of identifying and analysing the usage of Sinhala words in YouTube comments compared to their usage in the general domain. Understanding these differences can help highlight the richness of the Sinhala YouTube comment domain, identify rarely used words within YouTube comments that are common in the general domain, and pinpoint frequently used words within YouTube comments that are not as common in the general domain.

To construct the word corpus for the general domain, the authors employed a rigorous process. Authors utilized Sinhala Wikipedia articles, Sinhala news articles, and data from previous research studies on Sinhala NLP, as referenced in the survey paper by de Silva [1]. Sinhala Wikipedia articles were extracted by the authors under the article names "භූගෝල විද්‍යාව", "භූගෝලීය කලාප","සමාජ විද්‍යාව","කලා,ඉතිහාසය","භූවේදය","ගණිතය","කලාව","ඉතිහාසය" and "නවීන විද්‍යාව" in order to enrich the general domain word corpus. Further, the authors extracted data from 38 Sinhala newspaper articles from Divaina online newspaper and 44 Sinhala newspaper articles from Lankadeepa online newspaper ranging the articles from December 2023 to April 2024. Apart from this as aforementioned authors utilized available resources from previous research work including the Sinhala news extract used in the research done by de Silva [3], Jayawickrama et al. [25] and Wijayarathna and Upeksha [26] and also extract from government documents available for access created by Pathirennehelage et al. [27]. Then all these data sources were categorised into data from Newspapers, Wikipedia and Government Documents for analysis purposes.

The filtered words from each corpus, where filtration done in order to remove the non Sinhala words, symbols and numbers etc. preserving the zero-width-joiner were loaded and combined to form a unique set of words. To facilitate visualization, a one-hot encoding scheme was employed for each word, representing its presence in the respective sources. This encoding allowed for color-coding the words based on their source, making it easier to interpret the visualizations.

To put things in perspective, a Word2Vec model was trained on the combined corpus, which included words from all three sources. This training provided dense, vector representations for each word in the combined set.

In order to visualize these high-dimensional word embeddings, t-SNE (t-distributed Stochastic Neighbor Embedding) was used to reduce the dimensionality to two components. t-SNE is a powerful technique for visualizing high-dimensional data, making it suitable for this analysis of word vectors. The resulting 2D vectors from t-SNE were then plotted to observe the clustering and distribution patterns of the words. Separate plots were created for each combination of corpus sources illustrated in the Figures 5a to 5g. There is a larger cluster created from the words

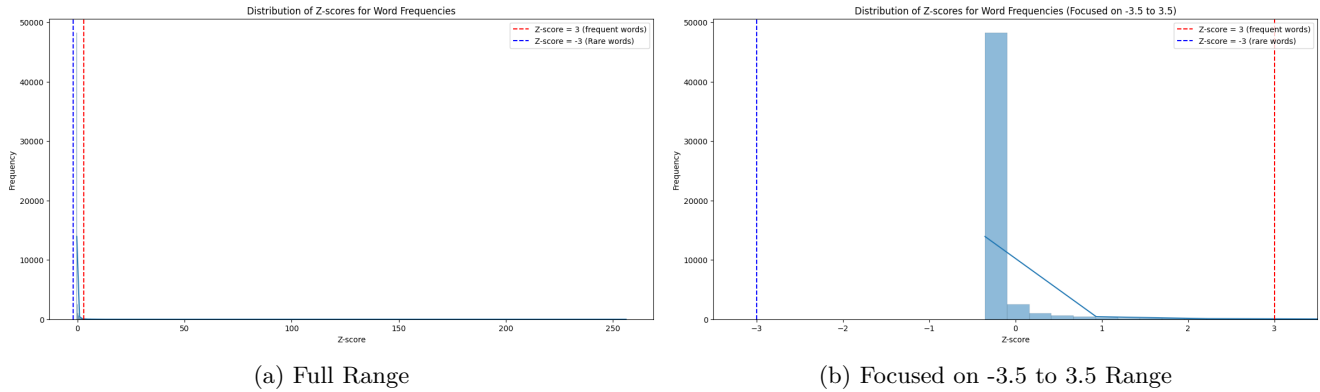

(a) Full Range

(b) Focused on -3.5 to 3.5 Range

Fig. 4: Distribution of Z Value of Word Frequency

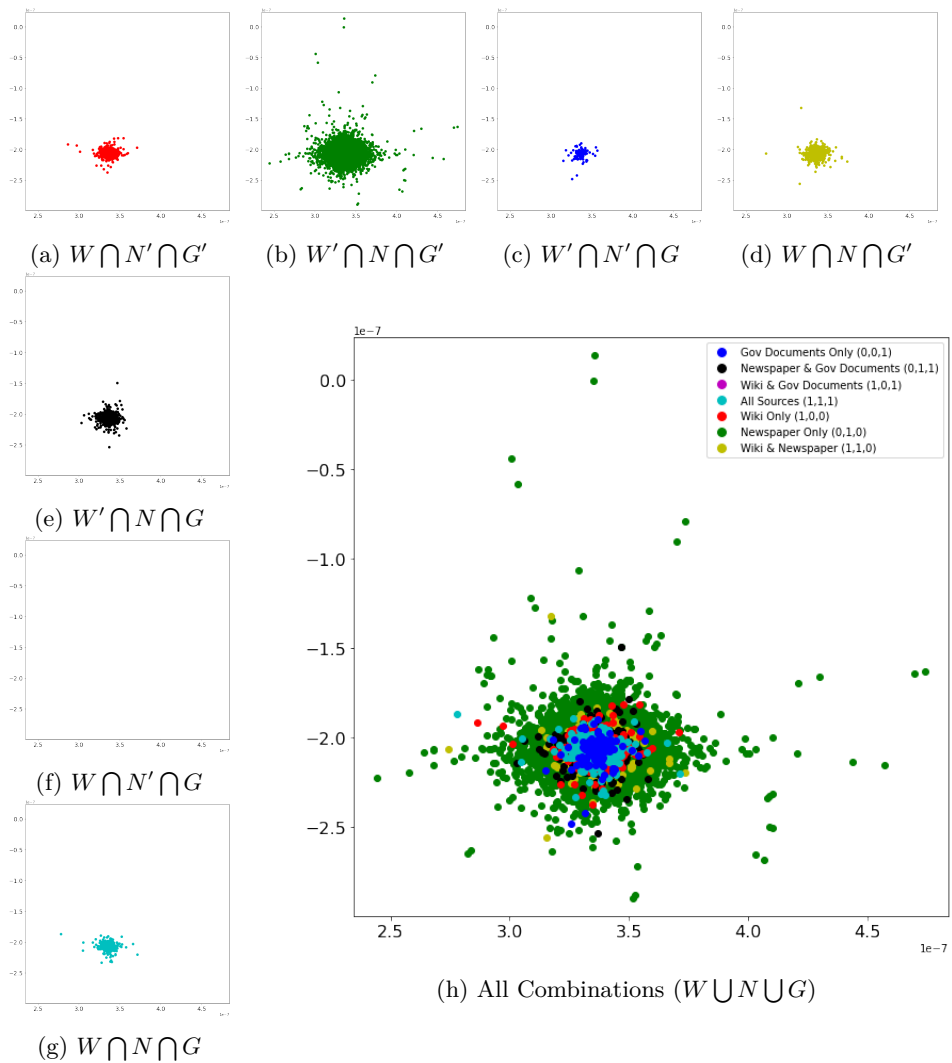

(a) $W \bigcap N' \bigcap G'$

(b) $W' \bigcap N \bigcap G'$

(c) $W' \bigcap N' \bigcap G$

(d) $W \bigcap N \bigcap G'$

(e) $W' \bigcap N \bigcap G$

(f) $W \bigcap N' \bigcap G$

(g) $W \bigcap N \bigcap G$

(h) All Combinations $(W \bigcup N \bigcup G)$

Fig. 5: Word Distribution among Corpora, where the set of words in each of the corpora are denoted as: $W=$ Wikipedia Corpus, $N=$ NEWS Corpus, $G=$ Government Document Corpus

only in News paper articles. Then the words from the intersection of Wikipedia and News paper, Government documents and News paper create the larger clusters respectively.

A comprehensive scatter plot was also created to visualize all the combinations together, offering a holistic view of the word embeddings shown in Figure 5. This combined plot allowed for the observation of how words from different source combinations are positioned relative to one another in the vector space, with a carefully designed legend for easy interpretation. Words appearing in multiple sources are more interspersed, indicating that shared context across sources leads to more similar embeddings. This intermixing highlights the semantic consistency of such words across different domains, as their meanings and usages are reinforced by their presence in multiple corpora.

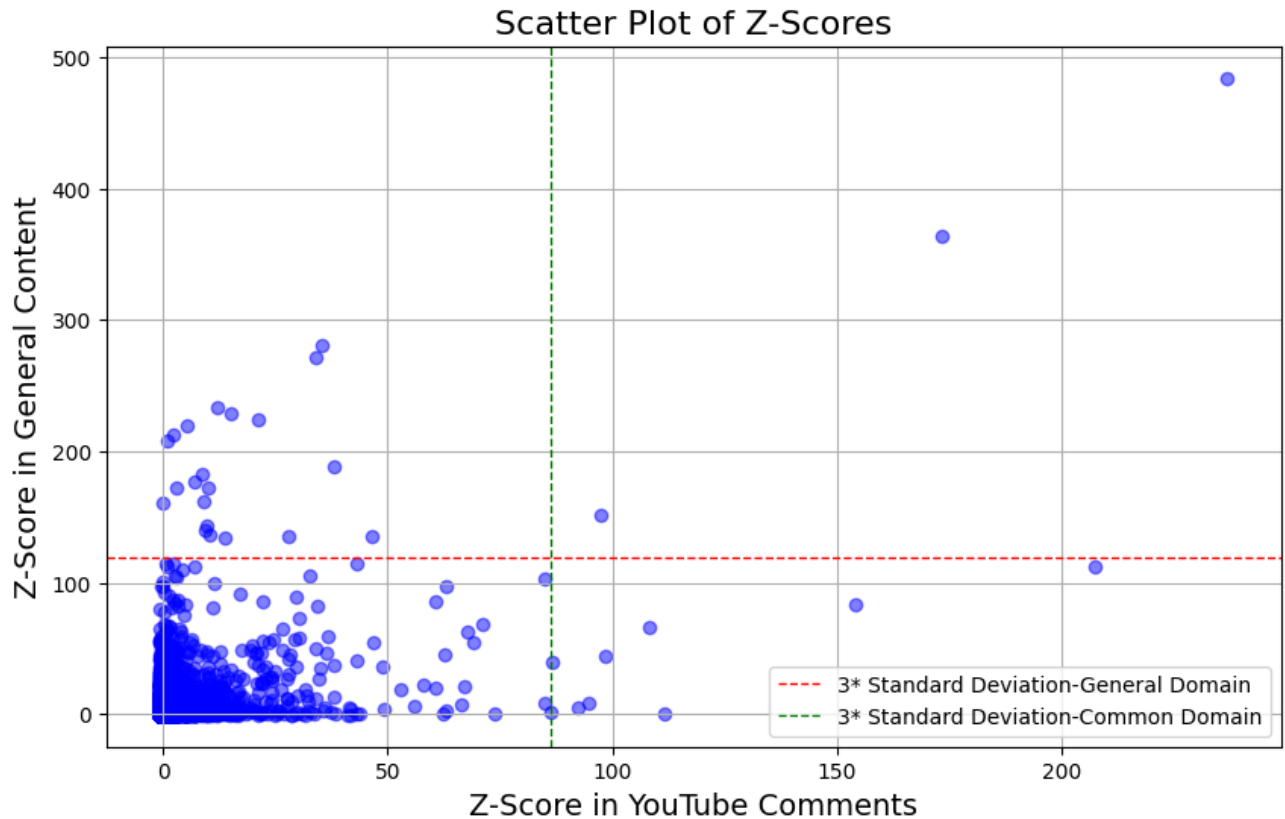

Fig. 6: Scatter Plot of Z-Scores

The authors utilized the same rigorous process to identify unique words in the YouTube comment domain for the general Sinhala domain data. It was found that there were 121,850 unique words in the general domain corpus, which is slightly more than twice the number of unique words in the YouTube comment domain, which had 54,834 unique words.

Subsequently, the authors identified 20,015 common words between the YouTube comment domain and the general domain. This indicates that 36% of the unique words found in the YouTube comment domain are also present in the general domain. This comparison underscores the lexical diversity between the two domains and highlights the overlap in vocabulary, providing valuable insights into the distinctive and shared linguistic features of Sinhala used in different domains.

Moreover, this analysis demonstrates that the YouTube comment domain serves as a representative subset of the general Sinhala language domain. The substantial overlap in unique words suggests that the linguistic patterns observed in YouTube comments can be reflective of the broader usage of Sinhala in various domains. This finding is significant for understanding the dynamic and evolving nature of the Sinhala language across different media and platforms.

Subsequent to this, the authors further analyzed the distribution of the commonly available words in both the YouTube comment domain and the general Sinhala domain. The Z-scores for the common words were calculated separately for each domain. The distribution of Z-scores for these common words in both the YouTube comment domain and the general domain is visualized in Figure 6. This analysis was divided into four quadrants using values corresponding to three times the standard deviation for both domains. This detailed examination provides a clearer understanding of how frequently common words are used across different domains, highlighting significant deviations and patterns. By employing Z-scores, the study effectively normalizes the data, allowing for a more precise comparison of word usage frequency between the YouTube comment domain and the general Sinhala domain.

Subject to the above analysis, it was observed that certain commonly available words, such as "වගේ", "සුපිරි", "එක", "එකක්", "නෑ", "මේක", "ලස්සන", and "ලස්සනයි", are significantly more frequent in the YouTube comment domain compared to the general Sinhala domain. This observation is based on a threshold of three times the standard deviation. This finding highlights the distinct linguistic trends within the YouTube comment domain, suggesting a unique usage pattern of certain Sinhala words that differs from the general language corpus. Such differences underscore the importance of domain-specific analyses in understanding language use in varying domains and can provide valuable insights for further research in Sinhala NLP.

Further analysis revealed that there were 21 words out of the common words that appeared significantly more frequently in the general Sinhala domain compared to the YouTube comment domain, based on the same threshold of three times the standard deviation. These words include: "ඇති", "කර", "බව", "ද", "අතර", "ලෙස", "සඳහා", "එම", "වන", "හා", "එය", "කළ", "මෙම", "වූ", "ඇත", "සහ", "ය", "කරන", "නිසා", "වී", and "අපි".

Additionally, it was observed that 19,981 words fell within the range where their frequency was less than three times the standard deviation for both domains. Notably, the words "ඒ", "මේ", and "නම්" were found in the range where the Z-score for both domains exceeded three times the standard deviation, which is represented in the top-right quadrant of Figure 6.

This detailed comparison highlights the unique and common linguistic features across different domains, providing deeper insights into the usage patterns of Sinhala words. Such findings are crucial for enhancing the understanding of language use in varied domains and can significantly contribute to the development of domain-specific natural language processing (NLP) tools for Sinhala.

## V. Conclusion

The exploration of Music Information Retrieval (MIR) and Music Emotion Recognition (MER) in the context of Sinhala songs presents an innovative and culturally significant contribution to the field. This research, situated at the intersection of MIR, MER, and Natural Language Processing (NLP), addresses the underexplored domain of Sinhala music. The unique linguistic and cultural characteristics of Sinhala songs pose both challenges and opportunities, which this study meticulously navigates through the analysis of social media comments on YouTube.

A comprehensive dataset of 63,471 Sinhala YouTube comments was developed, representing a wide array of songs and eras within Sinhala music. The rigorous preprocessing steps ensured the integrity and reliability of the data, enabling a thorough analysis of emotional expressions and linguistic patterns. The subsequent identification and comparison of frequently occurring words and word pairs in Sinhala YouTube comments against a general Sinhala corpus highlighted distinct linguistic trends within the online conversations surrounding Sinhala songs.

The findings underscore the lexical diversity and emotional resonance encapsulated in Sinhala YouTube comments. Notably, words such as "ලස්සනයි" and "සුපිරි" and word pairs like "කියන්න වචන" and "හරිම ලස්සනයි" reflect the aesthetic appreciation and emotional engagement of the audience. The significant overlap of 36% in unique words between the YouTube comment domain and the general Sinhala domain suggests that the linguistic patterns observed in the comments are reflective of broader Sinhala language usage.

Moreover, the derivation of stop words and the detailed comparison of word frequencies revealed distinct linguistic patterns unique to the YouTube comment domain. The identification of words that are significantly more frequent in YouTube comments or in the general Sinhala domain highlights the dynamic nature of language use in different contexts. These insights are crucial for advancing NLP applications tailored to Sinhala, particularly in the domains of sentiment analysis and cultural studies.

In conclusion, this research not only bridges the gap in the scholarly exploration of Sinhala music but also contributes valuable resources and methodologies for future studies in MIR, MER, and NLP for underrepresented languages. The integration of social media comments as primary data sources provides a unique lens into the spontaneous and authentic emotional reactions of listeners, enriching our understanding of the complex relationship between music, language, and emotion within the Sinhala cultural milieu. The robust dataset and analytical approaches established in this study pave the way for further research, fostering a deeper appreciation and understanding of the linguistic and emotional dimensions of Sinhala music.

## References

[1] N. de Silva, "Survey on publicly available sinhala natural language processing tools and research," *arXiv preprint arXiv:1906.02358*, 2019.

[2] D. Upeksha, C. Wijayarathna, M. Siriwardena, L. Lasandun, C. Wimalasuriya, N. De Silva, and G. Dias, "Implementing a corpus for sinhala language," in *Symposium on Language Technology for South Asia*, vol. 2015, 2015, p. 3.

[3] N. de Silva, "Sinhala text classification: observations from the perspective of a resource poor language," *ResearchGate*, 2015.

[4] F. Guzmán, P.-J. Chen, M. Ott, J. Pino, G. Lample, P. Koehn, V. Chaudhary, and M. Ranzato, "The flores evaluation datasets for low-resource machine translation: Nepali-english and sinhala-english," *arXiv preprint arXiv:1902.01382*, 2019.

[5] Y. Wijeratne and N. de Silva, "Sinhala language corpora and stopwords from a decade of sri lankan facebook," *arXiv preprint arXiv:2007.07884*, 2020.

[6] V. Dhananjaya, P. Demotte, S. Ranathunga, and S. Jayasena, "Bertifying sinhala–a comprehensive analysis of pre-trained language models for sinhala text classification," *arXiv preprint arXiv:2208.07864*, 2022.

[7] H. Hettiarachchi, D. Premasiri, L. Uyangodage, and T. Ranasinghe, "Nsina: A news corpus for sinhala," *arXiv preprint arXiv:2403.16571*, 2024.

[8] M. A. Casey, R. Veltkamp, M. Goto, M. Leman, C. Rhodes, and M. Slaney, "Content-based music information retrieval: Current directions and future

challenges," *Proceedings of the IEEE*, vol. 96, no. 4, pp. 668–696, 2008.
[9] F. Simonetta, S. Ntalampiras, and F. Avanzini, "Multimodal music information processing and retrieval: Survey and future challenges," in *2019 international workshop on multilayer music representation and processing (MMRP)*.  IEEE, 2019, pp. 10–18.
[10] S. Hizlisoy, S. Yildirim, and Z. Tufekci, "Music emotion recognition using convolutional long short term memory deep neural networks," *Engineering Science and Technology, an International Journal*, vol. 24, no. 3, pp. 760–767, 2021.
[11] Y. E. Kim, E. M. Schmidt, R. Migneco, B. G. Morton, P. Richardson, J. Scott, J. A. Speck, and D. Turnbull, "Music emotion recognition: A state of the art review," in *Proc. ismir*, vol. 86, 2010, pp. 937–952.
[12] Y.-H. Yang and H. H. Chen, "Machine recognition of music emotion: A review," *ACM Transactions on Intelligent Systems and Technology (TIST)*, vol. 3, no. 3, pp. 1–30, 2012.
[13] S. Koelsch, "Investigating emotion with music: neuroscientific approaches," *Annals of the New York Academy of Sciences*, vol. 1060, no. 1, pp. 412–418, 2005.
[14] M. V. Thoma, S. Ryf, C. Mohiyeddini, U. Ehlert, and U. M. Nater, "Emotion regulation through listening to music in everyday situations," *Cognition & emotion*, vol. 26, no. 3, pp. 550–560, 2012.
[15] Y. S. Can, B. Mahesh, and E. André, "Approaches, applications, and challenges in physiological emotion recognition—a tutorial overview," *Proceedings of the IEEE*, 2023.
[16] S. Lai, L. Xu, K. Liu, and J. Zhao, "Recurrent convolutional neural networks for text classification," in *Proceedings of the AAAI conference on artificial intelligence*, vol. 29, no. 1, 2015.
[17] L.-C. Yu, J. Wang, K. R. Lai, and X. Zhang, "Refining word embeddings using intensity scores for sentiment analysis," *IEEE/ACM Transactions on Audio, Speech, and Language Processing*, vol. 26, no. 3, pp. 671–681, 2017.
[18] T. Mikolov, K. Chen, G. Corrado, and J. Dean, "Efficient estimation of word representations in vector space," *arXiv preprint arXiv:1301.3781*, 2013.
[19] T. Mikolov, I. Sutskever, K. Chen, G. S. Corrado, and J. Dean, "Distributed representations of words and phrases and their compositionality," *Advances in neural information processing systems*, vol. 26, 2013.
[20] P. Donnelly and A. Beery, "Evaluating large-language models for dimensional music emotion prediction from social media discourse," in *Proceedings of the 5th International Conference on Natural Language and Speech Processing (ICNLSP 2022)*, 2022, pp. 242–250.
[21] P. Sarakit, T. Theeramunkong *et al.*, "A music video recommender system based on emotion classification on user comments," *Doctoral dissertation*, 2015.
[22] Y. Agrawal, R. G. R. Shanker, and V. Alluri, "Transformer-based approach towards music emotion recognition from lyrics," in *European conference on information retrieval*.  Springer, 2021, pp. 167–175.
[23] E. Çano, "Text-based sentiment analysis and music emotion recognition," *arXiv preprint arXiv:1810.03031*, 2018.
[24] D. Lakmal, S. Ranathunga, S. Peramuna, and I. Herath, "Word embedding evaluation for sinhala," in *Proceedings of the Twelfth Language Resources and Evaluation Conference*, 2020, pp. 1874–1881.
[25] V. Jayawickrama, A. Ranasinghe, D. C. Attanayake, and Y. Wijeratne, "A corpus and machine learning models for fake news classification in sinhala," 2021.
[26] C. Wijayarathna and D. Upeksha, "Sinmin-sinhala corpus project."
[27] R. Pathirennehelage, N. Ihalapathirana, A. Mohamed, M. Ranathunga, S. Jayasena, S. Dias *et al.*, "Automatic creation of a sentence aligned sinhala-tamil parallel corpus," in *Proceedings of the 6th Workshop on South and Southeast Asian Natural Language Processing (WAssanlp)*, 2016, pp. 124–132.